\documentclass[sigconf,nonacm]{acmart}
\AtBeginDocument{%
  \providecommand\BibTeX{{%
    \normalfont B\kern-0.5em{\scshape i\kern-0.25em b}\kern-0.8em\TeX}}}

\begin{document}

\title{Room Classification on Floor Plan Graphs using Graph Neural Networks}

\author{Abhishek Paudel}
\affiliation{%
  \institution{George Mason University}
  \city{Fairfax}
  \state{Virginia}
  \country{USA}
}
% \email{apaudel4@gmu.edu}

\author{Roshan Dhakal}
\affiliation{%
  \institution{George Mason University}
  \city{Fairfax}
  \state{Virginia}
  \country{USA}
 }
% \email{rdhakal2@gmu.edu}

\author{Sakshat Bhattarai}
\affiliation{%
  \institution{George Mason University}
  \city{Fairfax}
  \state{Virginia}
  \country{USA}
 }
% \email{sbhatta8@gmu.edu}


\begin{abstract}
We present our approach to improve room classification task on floor plan maps of buildings by representing floor plans as undirected graphs and leveraging graph neural networks to predict the room categories. Rooms in the floor plans are represented as nodes in the graph with edges representing their adjacency in the map. We experiment with House-GAN dataset that consists of floor plan maps in vector format and train multilayer perceptron and graph neural networks. Our results show that graph neural networks, specifically GraphSAGE and Topology Adaptive GCN were able to achieve accuracy of 80\% and 81\% respectively outperforming baseline multilayer perceptron by more than 15\% margin.

\end{abstract}


\maketitle
\pagestyle{plain} % removes running headers

\section{Introduction}
A floor plan is a map representing the layout of a floor in a building. Floor plans generally illustrate the locations of walls, doors windows and other structural properties in the floor. Although it has its canonical usage in architectural design, various types of maps resembling a floor plan is widely used in the context of indoor navigation by mobile robots. Robotic mapping techniques like SLAM are commonly leveraged to construct a map of floor of a building which the robot can then use for localization and navigation. Such generated maps contain no explicit information about room usage. However, information about room usage can be useful for robot navigation. In this paper, we present our approach to improve room classification task on floor plan maps of buildings by representing floor plans as undirected graphs and leveraging graph neural networks to predict the room categories. The key motivation towards this approach lies in the intuition that category of a room has some predictable relationship with the neighboring rooms. This relationship can be captured using graph neural networks that take into account the influence of neighboring nodes in the graph through message passing mechanism between nodes.

We generate room connectivity graph from floor plans in House-GAN dataset by representing each room as nodes in the graph with edges between them representing their adjacency. We experiment with the dataset using four types of graph neural networks: graph convolutional networks (GCN)\cite{kipf2016semisupervised}, graph attention network (GAT)\cite{velikovi2017graph}, GraphSAGE\cite{hamilton2017inductive} and topology adaptive graph convolutional networks (TAGCN)\cite{du2017topology}, along with a baseline multi-layer perceptron (MLP). Our experiments demonstrate that GraphSAGE and TAGCN significantly outperform all other models in terms of classification accuracy. Our experiments with deepening the layers in the network also show that GraphSAGE and TAGCN do not underperform upon deepening the layers in the network in contrast to what one would generally expect due to oversmoothing\cite{li2019deepgcns}.

\begin{figure}
\centering
\includegraphics[width=8cm]{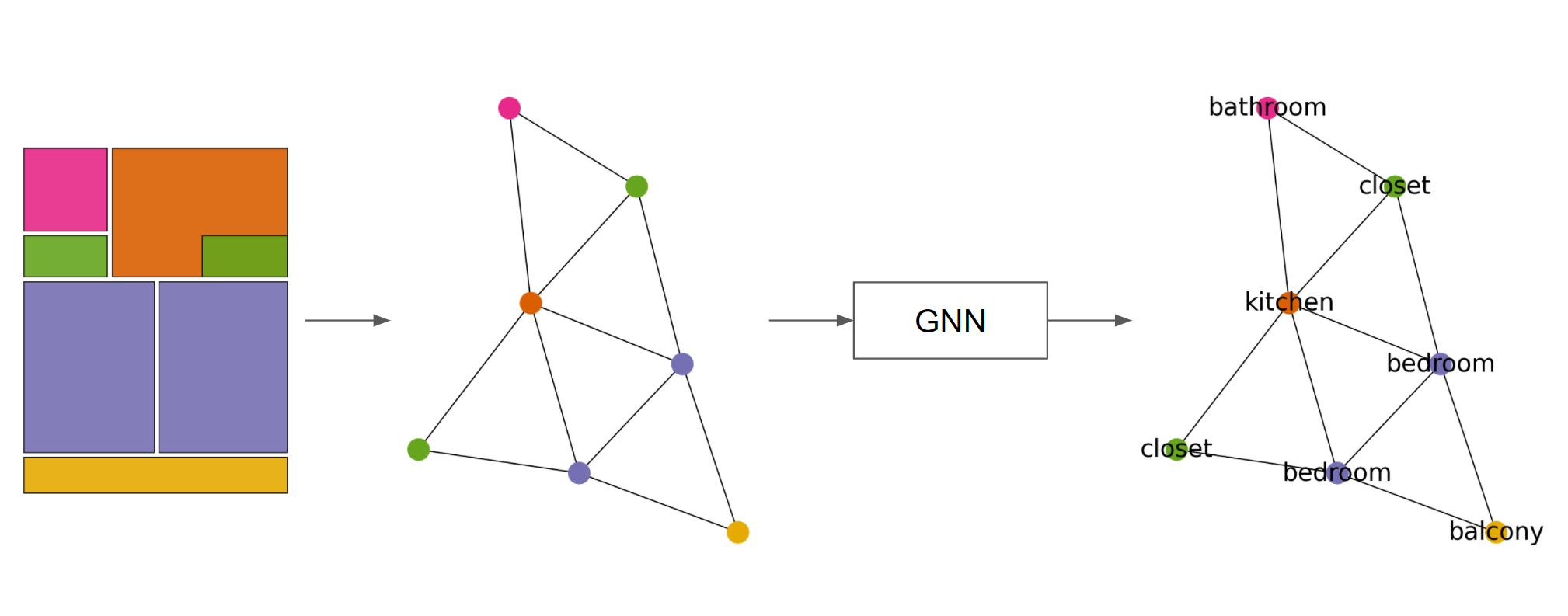}
\caption{Floor plan map can be represented as an undirected graph with rooms as nodes, and edges representing their adjacency in the map. A graph neural network then takes this graph as input to make prediction of room categories as a node classification task.}
\label{fig:approach}
\end{figure}

\section{Related Work}
Detecting and classifying elements of floor plans have been studied as a part of architectural floor plan analysis for many years using various rule-based heuristics approaches and image processing approaches\cite{ahmed2012automatic}\cite{dominguez2012semiautomatic}\cite{dosch2000complete}\cite{mace2010system}\cite{barducci2012object}. Later, machine learning approaches that use fully connected neural networks and convolutional neural networks were also used for tasks like vectorization, segmentation and detection \cite{liu2017raster}\cite{de2014statistical}\cite{dodge2017parsing}\cite{zeng2019deep}. In \cite{barducci2012object} and \cite{dominguez2012semiautomatic}, elements in floor plans are represented as a graph structure for analysis and recognition of floor plan. Most of these aforementioned tasks have focused on semantic understanding of floor plan in the context of architectural design.

With the introduction of graph neural networks (GNNs) in the recent years, a new frontier of learning over graph structures has been opened. In \cite{hu2021roomsemantics}, the authors use relational graph convolutional network (RGCN) in addition to traditional machine learning to infer the usage of rooms in public buildings using floor plan represented as graph. However, their results show that RGCN does not outperform traditional machine learning approaches like random forest.

In the field of robotics, a robotic agent uses a map of the environment for localization and navigation. These maps are broadly classified into two categories: metric maps and topological maps\cite{thrun1998learning}. Metric maps like occupancy grid maps\cite{moravec1985high} represent the environment in 2D grid distinguishing obstacles like walls and objects from free space where robot can navigate. Topological maps represent the environment as graph where nodes represent notable places and edges represent navigability between them generally without metric information. 
In this context, making inferences about unknown spaces is an active field of research in robotic navigation. In \cite{liu2014extracting}, a method is proposed to infer semantic information from occupancy grid map represented as a scene graph. In \cite{aydemir2012can}, the authors propose to predict the topology and the categories of rooms given a partial map. The map of the environment is represented as a graph with rooms as nodes and their connectivity as edges. Our work aims to use graph neural networks to demonstrate that inference on such graph representations can be improved with the recent advancements in learning over graph structures.

\section{Dataset}
We experiment with House-GAN dataset\cite{nauata2020housegan} that consists of 143,184 floor plans extracted in vector format from real floor plan images in Lifull Home's dataset\cite{lifullhomedataset}. There are 8 categories of rooms\footnote{The original size of the dataset and number of classes are slightly different. This is because we perform some data cleaning and also ignore irrelevant samples like those consisting of only one room in floor plan which would result in empty edges when represented as a graph.}.

The dataset consists of the following.
\begin{itemize}
    \item Room bounding boxes
    \item Room categories
    \item Wall coordinates
    \item Wall to room mapping
    \item Presence of door in walls
    \item Name of the original RGB image from Lifull Home's dataset
\end{itemize}

\begin{figure}
    \centering
    \includegraphics[width=8cm]{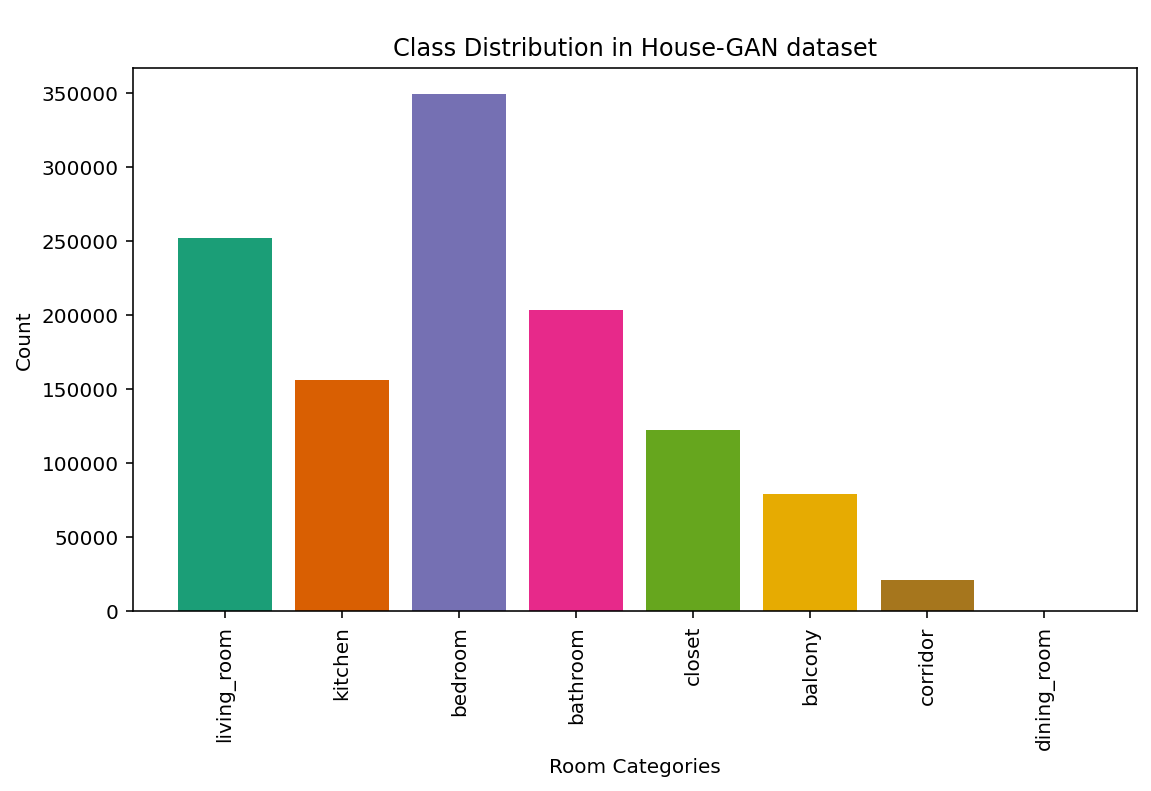}
    \caption{Class distribution in House-GAN dataset}
    \label{fig:class_distribution}
\end{figure}

\begin{figure}
    \centering
    \includegraphics[width=8cm]{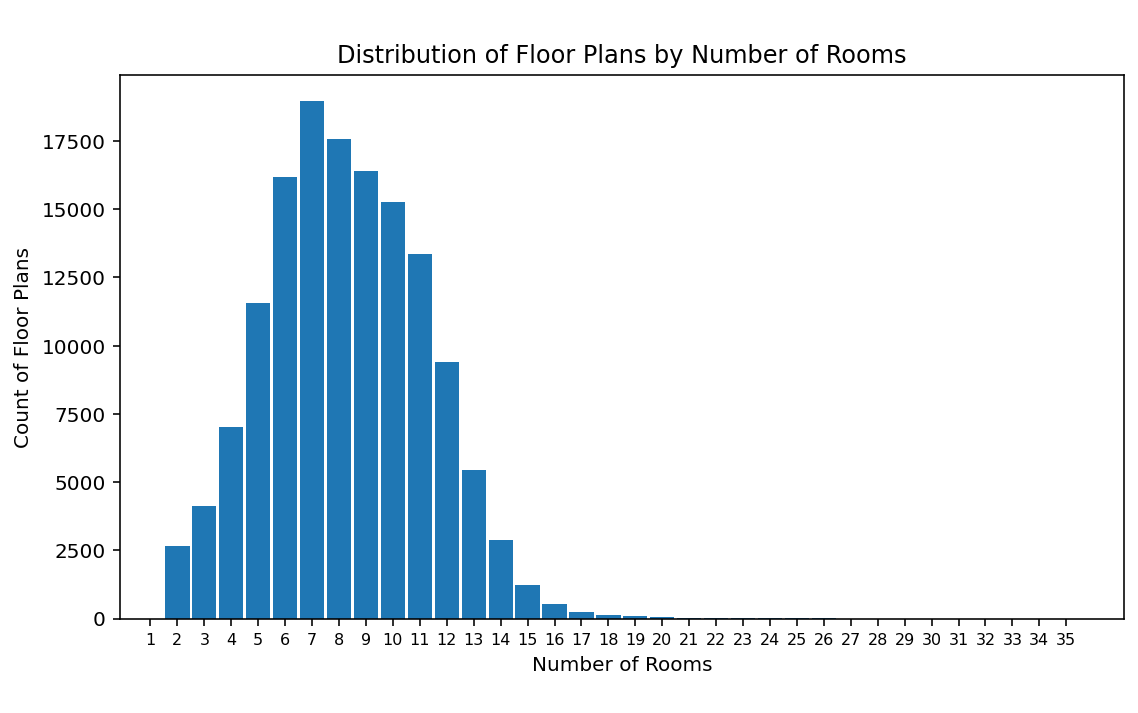}
    \caption{Distribution of floor plans by number of rooms}
    \label{fig:floor_plan_dist}
\end{figure}

\section{Approach}
Our approach consists of three stages. First, we generate undirected graphs from the vectorized floor plan maps in the House-GAN dataset. Second, we extract relevant node features from the geometric and structural properties of rooms that are available in the floor plans. Finally, we train multiple graph neural networks on the floor plan graphs in inductive settings as node classification task to predict the categories of the rooms.

\subsection{Graph Generation}
We generate the undirected graphs from vectorized floor plan maps with nodes in the graph representing the rooms, and edges representing the adjacency between two rooms. We consider two rooms to be adjacent if the distance between them is smaller than a certain threshold. This is done on the normalized bounding boxes of rooms with the threshold value set to 0.03 meaning that two rooms are adjacent if they are within 3\% of the length of floor plan map.

\subsection{Node Feature Extraction}
We extract the following node features corresponding to each room in the floor plan.

\begin{itemize}
    \item Area of the room
    \item Length of the room
    \item Width of the room
    \item Number of doors in the room
    \item Whether a room is a parent room
    \item Whether a room is a child room
\end{itemize}

The first four features: area, length, width, and number of doors in the list above are self-explanatory. The last two features in the list above are useful for nested rooms i.e. room within a room. This is because some rooms in the floor plan are inside or intersect with another room and could be useful as a distinguishing feature, for example the closet could be inside a bedroom, the dining room could be inside the kitchen, and so on. If the intersection of a room with another room is greater than 70\% of its area, we assume that room to be a child room and the another room that it intersects with to be a parent room. These are both binary features with two possible values: 0 or 1.

\subsection{Models}
We experiment with multilayer perceptron (MLP) as a baseline model, and graph neural network (GNN) models. The multilayer perceptron model only takes node features as input without any graph structures that define connectivity between nodes, while the graph neural network models take both node features and the graph structures represented as a list of edges between nodes as input. For graph neural network models, we experiment with the following four variants.

\begin{itemize}
    \item \textbf{Graph Convolutional Network}
    
        Proposed by Kipf et. al.\cite{kipf2016semisupervised}, Graph Convolutional Network (GCN) is one of the first models that proposed a general approach for convolution on graphs in transductive settings. Intuitively, this method aggregates features from neighboring nodes after a linear transformation to obtain new features for the node.

    \item \textbf{Graph Attention Network}
    
        Proposed by Veličković et. al.\cite{velikovi2017graph}, Graph Attention Network (GAT) improves graph convolutions by leveraging masked self-attentional layers to address the shortcomings of prior methods and thus enabling the model to learn different weights for different nodes in a neighborhood.

    \item \textbf{GraphSAGE}
    
        Proposed by Hamilton et. al.\cite{hamilton2017inductive}, GraphSAGE extends previous approaches for graph convolution to inductive settings by learning to generate node embeddings by sampling and aggregating features from the node's local neighborhood.

    \item \textbf{Topology Adaptive Graph Convolutional Network}
    
        Proposed by Du et. al.\cite{du2017topology}, Topology Adaptive Graph Convolutional Network (TAGCN) extends previous approaches by learning a fixed-size filter to perform convolution on graphs that resembles convolutions on images. It does not approximate convolution, unlike previous methods.

\end{itemize}

\begin{figure}
    \centering
    \includegraphics[width=5cm]{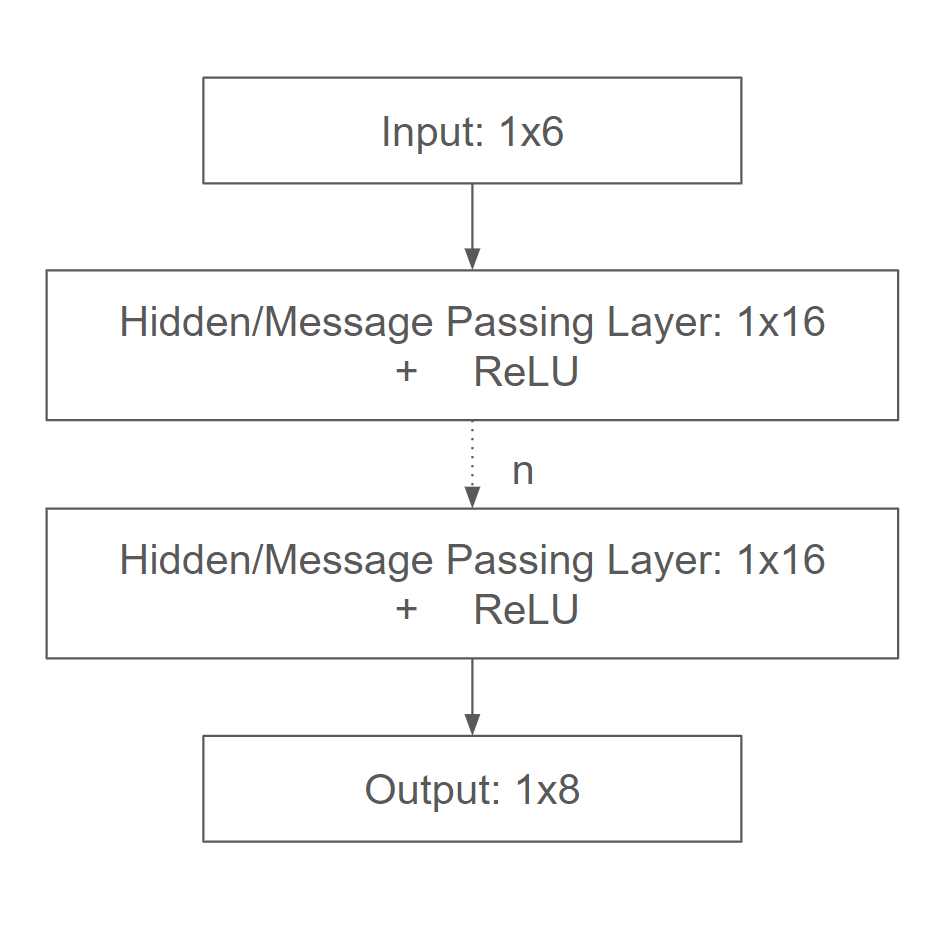}
    \caption{General Model Architecture}
    \label{fig:architecture}
\end{figure}

We implement a general model architecture as shown in Figure \ref{fig:architecture} for MLP and all GNN models for a fair comparison between them. The input layer takes in a vector of size 6 corresponding to 6 node features. We experiment with varying number of hidden or message passing layers each with a size of 16 and followed by ReLU activation function. The output layer has a size of 8 corresponding to 8 categories of rooms.

\section{Experimental Setup}
We split the House-GAN dataset into two parts for training and testing. We take 120,000 floor plans for training and the remaining 23,184 floor plans for testing. Preliminary small-scale experiments were run on a small fraction of the dataset for hyperparameter tuning before training on the full dataset. For large-scale training on full dataset, we use the following setup and hyperparameters for training.

\begin{itemize}
    \item GPU: NVIDIA Tesla K80 with 8GB memory
    \item Number of training epochs: 100
    \item Loss function: Cross-entropy loss
    \item Gradient descent optimizer: Adam
    \item Batch size: 128
    \item Learning rate: 0.004
    \item Learning rate scheduling: Exponential with step size of 10 and gamma of 0.8
\end{itemize}

We use PyTorch and PyTorch Geometric for our experiments. Large-scale experiments were run on ARGO, a research computing cluster provided by the Office of Research Computing at George Mason University. The cumulative training time to run all experiments with all models was approximately 75 hours.

\section{Results}
The best results obtained in terms of classification accuracy for each model are shown in Table \ref{tab:accuracy}. The best performance is achieved by TAGCN with 81.07\% test accuracy followed by GraphSAGE with test accuracy of 80.26\%. We observe that GCN and GAT had very low test accuracy of 54.85\% and 52.50\% respectively, and performed even worse than MLP which had the test accuracy of 65.74\%. It should also be noted that although we have not used any explicit regularization techniques in our models, the results demonstrate that overfitting is not an issue since all the models seem to have generalized well to the test set.

\begin{table}
    \caption{Classification Accuracy}
    \label{tab:accuracy}
    \begin{tabular}{ccc}
    \toprule
    Model & Train Accuracy & Test Accuracy \\
    \midrule
    MLP & 0.6585 & 0.6574 \\
    GCN & 0.5469 & 0.5485 \\
    GAT & 0.5269 & 0.5250 \\
    GraphSAGE & 0.8068 & 0.8026 \\
    TAGCN & 0.8140 & 0.8107 \\
    \bottomrule
    \end{tabular}
\end{table}

\begin{figure}
    \centering
    \includegraphics[width=8cm]{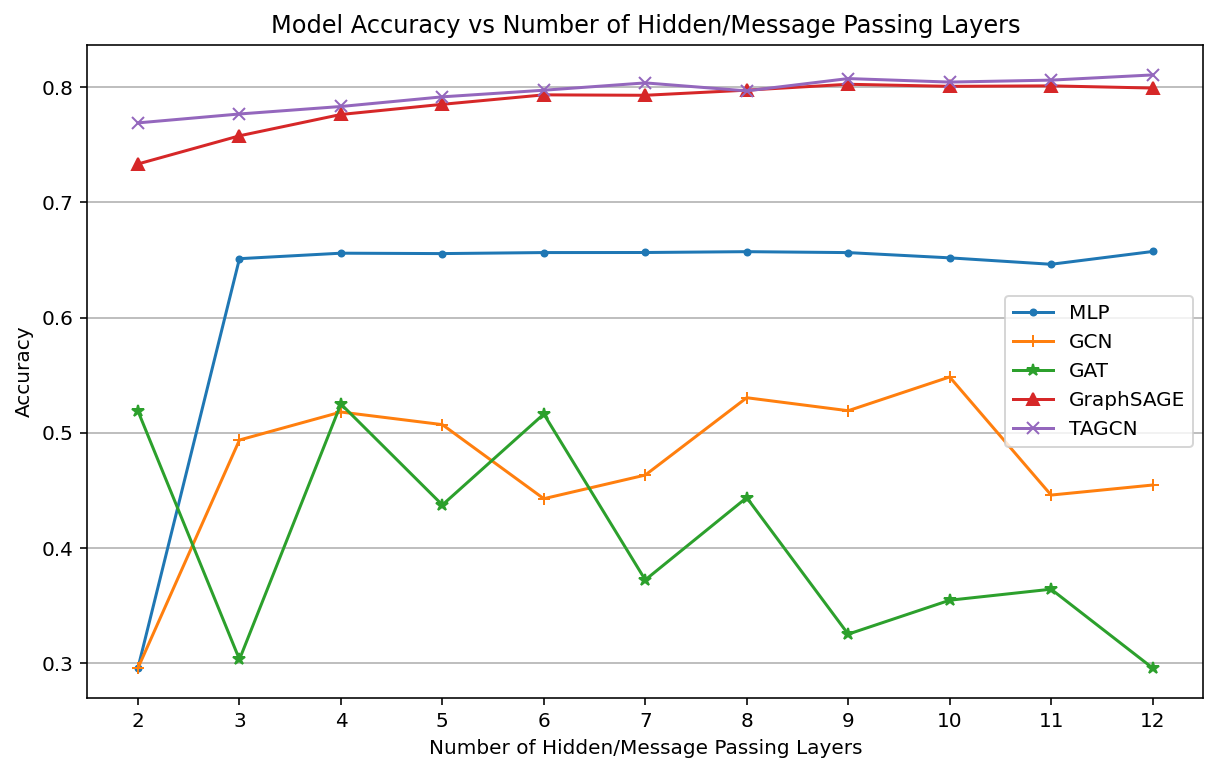}
    \caption{Visualization of test accuracy for each model with respect to the number of hidden/message passing layers}
    \label{fig:accuracy}
\end{figure}

To study the effects of oversmoothing which is common in graph neural networks, we train our models with a varying number of hidden/message passing layers ranging from 2 to 12 layers. The results are shown in Figure \ref{fig:accuracy}. We observe that MLP has no improvement in accuracy after 3 hidden layers. GraphSAGE and TAGCN have considerable improvements early on with only slight improvements as the number of layers increase. However, we do not see GCN and GAT perform better with more message passing layers. In fact, the performance of GAT generally keeps decreasing upon increasing the number of message passing layers which suggests that GAT might be more susceptible to oversmoothing. Surprisingly, both GraphSAGE and TAGCN have no decrease in performance upon increasing the number of message passing layers. This suggests that GraphSAGE and TAGCN are more robust to oversmoothing than other models.

\begin{figure}
    \centering
    \includegraphics[width=8cm]{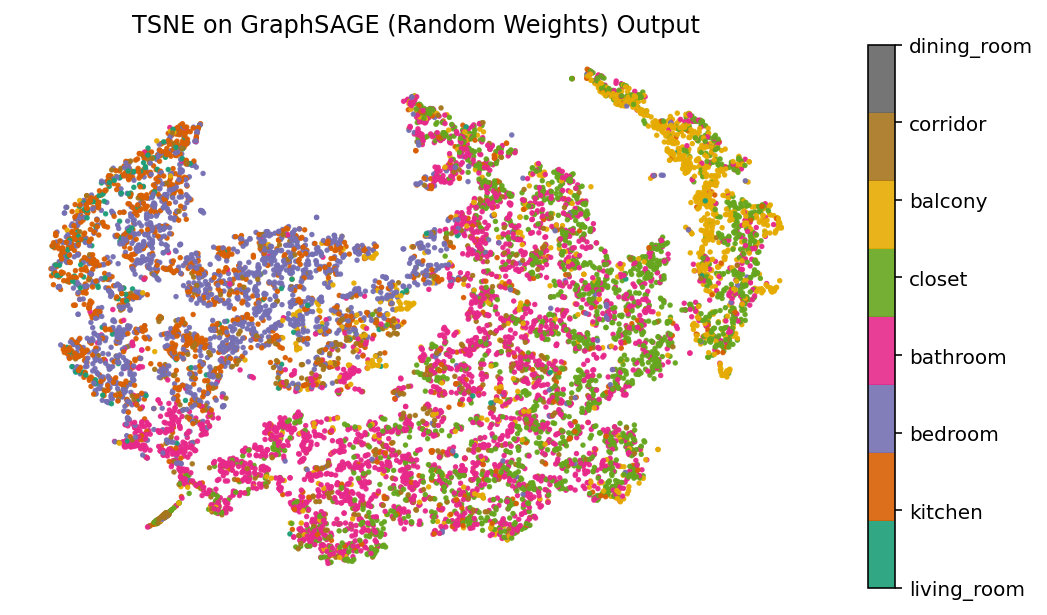}
    \caption{t-SNE on node embeddings obtained from GraphSAGE model initialized with random weights. Partial clusters for some classes are distinguishable.}
    \label{fig:tsne_graphsage}
\end{figure}

\begin{figure}
    \centering
    \includegraphics[width=8cm]{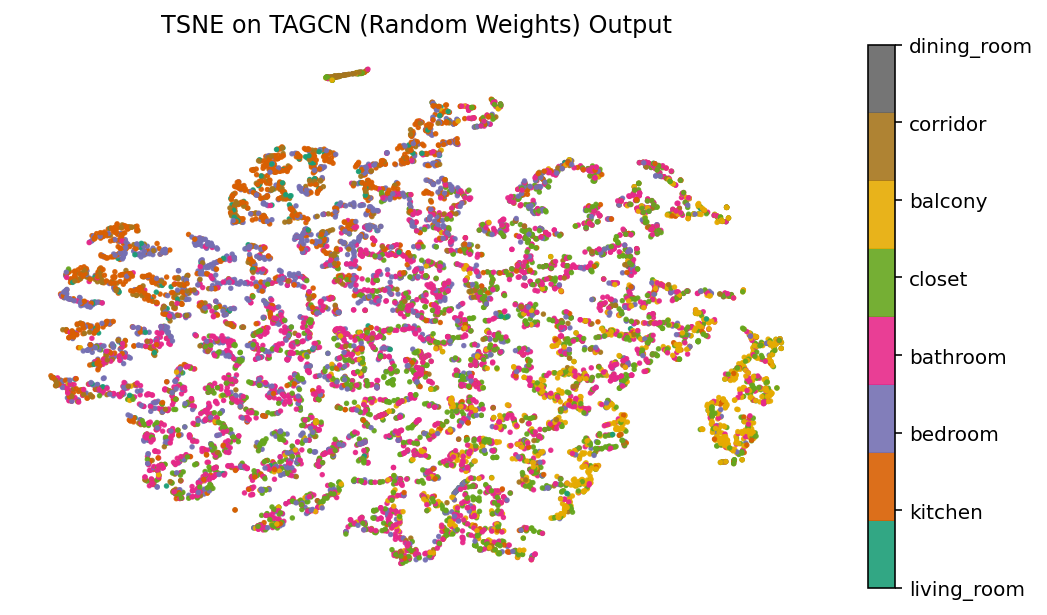}
    \caption{t-SNE on node embeddings obtained from TAGCN model initialized with random weights. Partial clusters for some classes are distinguishable.}
    \label{fig:tsne_tagcn}
\end{figure}

To understand why GCN and GAT were not able achieve a good performance unlike TAGCN and GraphSAGE both of which had superior performance, we study the characteristics of these models in terms of their abilities to capture relational inductive biases that are inherent to the graph structures due to the edges in the graph affecting how node features interact during message passing in the graph neural network\cite{battaglia2018relational}. To study this, we instantiate the GNN models with random weights and obtain the node embeddings for 10,000 samples of floor plan graphs from test split by propagating them through the untrained models. Then, we perform non-linear dimensionality reduction on the obtained node embeddings using t-Distributed Stochastic Neighbor Embedding (t-SNE). We observe that GraphSAGE and TAGCN were able to partially capture relational inductive biases even before training the network as demonstrated in Figure \ref{fig:tsne_graphsage} and \ref{fig:tsne_tagcn}. In both of the figures, we can see partial clusters for some of the classes in the dataset. For example, bedroom, bathroom and balcony have partially distinguishable clusters. We did not observe any meaningful clusters with GCN and GAT suggesting that these models were not quite capable of capturing relational inductive biases in the dataset. Trivially, no meaningful clusters were observed with MLP or with original features as one would generally expect strengthening our hypothesis that GraphSAGE and TAGCN are better at capturing relational inductive biases in the dataset resulting in better performance after training.

\section{Conclusion and Future Work}
Overall, our experiments demonstrate that graph neural networks can be leveraged to improve performance on room classification using floor plans represented as undirected graphs. Specifically, we observe that TAGCN and GraphSAGE show superior performance compared to other models like GCN, GAT and MLP we experimented with for this task. It was a surprising finding that GCN and GAT, which leveraged graph structures could not even perform on par with MLP which did not leverage graph structures. Additionally, our experiments to study relational inductive bias on untrained graph neural networks suggest that GCN and GAT are likely not very capable for capturing relations between the nodes in the graph resulting in subpar performance. However, further experiments would be needed to confirm this hypothesis. We also observe decreased performance of GAT as number of message passing layers were increased while GCN only has decreased performance when number of layers is 11 or 12. These results are likely due to oversmoothing. However, it was intriguing that GraphSAGE and TAGCN had either improved or similar performance as more layers were stacked. This suggests that GraphSAGE and TAGCN are more robust to oversmoothing than the other models. And again, more experiments would be needed to conclusively assert this statement.

In terms of improving the performance over our current approach, it would be interesting to explore whether adding more node features could be useful. One could also eliminate the need for manual feature extraction by using CNN. One possible direction for this could be to use the image of the floor plan with specific room segmented as node property and leveraging CNN to extract features from these images to be used as node features. This would however be computationally more expensive than our current approach.

\begin{acks}
All experiments were run on ARGO, a research computing cluster provided by the Office of Research Computing at George Mason University. We would like to thank the Office of Research Computing for providing the computing infrastructure to run these experiments.
\end{acks}

\bibliographystyle{ACM-Reference-Format}
\bibliography{main}

% \appendix


\end{document}